\documentclass[journal]{IEEEtran}
\usepackage{amsfonts,amssymb}
\usepackage{graphicx}
\usepackage{amsmath}
\usepackage{cases}
\ifCLASSINFOpdf
\else
\fi

\begin{document}
%
\title{ResMGCN: Residual Message Graph Convolution Network for Fast Biomedical Interactions Discovering}
%
%
%
\author{
  \IEEEauthorblockN{
      Zecheng Yin\\
      }
\IEEEauthorblockA{
    The Chinese University of Hong Kong (Shenzhen), Shenzhen, China\\
}
\IEEEauthorblockA{yinzecheng.cuhk@gmail.com}
}

%
%

\markboth{Journal of \LaTeX\ Class Files,~Vol.~14, No.~8, August~2015}%
{Shell \MakeLowercase{\textit{et al.}}: Bare Demo of IEEEtran.cls for IEEE Journals}
%



\maketitle

\begin{abstract}
Biomedical information graphs are crucial for interaction discovering of biomedical information in modern age, such as identification of multifarious molecular interactions and drug discovery, which attracts increasing interests in biomedicine, bioinformatics, and human healthcare communities.  Nowadays, more and more graph neural networks have been proposed to learn the entities of biomedical information and precisely reveal biomedical molecule interactions with state-of-the-art results. These methods remedy the fading of features from a far distance but suffer from remedying such problem at the expensive cost of redundant memory and time. In our paper, we propose a novel Residual Message Graph Convolution Network (ResMGCN) for fast and precise biomedical interaction prediction in a different idea. Specifically, instead of enhancing the message from far nodes, ResMGCN aggregates lower-order information with the next round higher information to guide the node update to obtain a more meaningful node representation. ResMGCN is able to perceive and preserve various messages from the previous layer and high-order information in the current layer with least memory and time cost to obtain informative representations of biomedical entities. We conduct experiments on four biomedical interaction network datasets, including protein-protein, drug-drug, drug-target, and gene-disease interactions, which demonstrates that ResMGCN outperforms previous state-of-the-art models while achieving superb effectiveness on both storage and time. Our code and involved datasets are publicly available at https://github.com/Yonggie/ResMGCN.
\end{abstract}


%
\IEEEpeerreviewmaketitle

\section{Introduction}
%
%
%
%
A biological information system is an amazingly sophisticated and interesting network consisting of various biological molecules such as proteins, genes, as well as the interaction/reactions between these entities. The bioinformation between these molecules can be represented as a information graph, with molecular entities as nodes and interactions as edges between them. Distinguished from language or image, graph structure representation of the system offers a conceptual and intuitive precondition for further explore to understand or discover the direct or indirect interactions between molecular entities in the biological system\cite{system_level}\cite{biosnapnets}. Such technology boosted the explore of biology in system level and made novel discoveries in interactions such as drug-drug interactions (DDI)\cite{ddi}, drug-target interactions(DTI)\cite{dti}, gene-disease associations (GDI)\cite{gdi}, and protein-protein interactions (PPI)\cite{ppi}.

Recently, a great number of deep learning methods are proposed to achieve promising performance across various domains, such as recommendation systems\cite{smgcn}\cite{gnn_survey}, chemistry\cite{chemi}, citation networks\cite{dmgi}\cite{hdgi}\cite{dgi}, social networks\cite{social} and medicine\cite{equibind}. Most of these works are based on Graph Convolution Network (GCN)\cite{gcn}. GCN is a popular work that passes the messages of neigbhouring nodes to the central node repeatedly by stacking multiple GCN layers, to learn the predictive representation of the nodes for downstream tasks. Although GCN-based models show great success in link prediction in bioinformatics\cite{ddi}\cite{bio}, there is a fundamental issue existing in GCN, which might depress the representation ability of biomedical graphs. GCN could not obtain information from different hierarchy at the same time and use these information to guide node updates. \cite{skipgnn} and HOGCN\cite{hogcn} and SkipGNN\cite{skipgnn} deem that GCN only considers the instant neighbours (also known as 1-hop neighbours) and message from nodes at a long distance is diminished on the way. Therefore, they conduct different methods to alleviate this problem. HOGCN\cite{hogcn} calculates messages from neighbours up to K hops at one time in one layer to combine information of neighbours from different distances, but it is computationally expensive and causes slow network propagation and optimization. SkipGNN\cite{skipgnn} leverages a copy of the original graph but with 1-hop neighbour removed and 2-hop neighbour connected to preserve information of 2-hop neighbours, and shows improvement. However, reconstructing a copy of the original graph and processing a new 2-hop connected graph itself is expensive in both storage and time.

Targeting this pain point, inspired by ResNet\cite{resnet} in computer vision, instead of handling the messages from nodes far away in expensive computational or storage cost, we propose Residual Message Graph Convolution Network (ResMGCN), a simple but effective improvement on GCN, a novel graph convolution that is able to effectively preserve lower order information and process together with the instant neighbour message to guide this layer's node update to obtain informative representation. We applied ResMGCN on an end2end biomedical entity representation learning that learns every entity of a biomedical system via an interaction network $\mathcal{G}$ and its features $X$ and conduct link prediction on two certain entities to determine whether they have meaningful interaction or not. Specifically, we first encode the biomedical graph using ResMGCN  that utilizes low-order and high-order information, then define a simple linear decoder to predict interactions between the nodes.

We demonstrate our ResMGCN's performance with state-of-the-art heuristic learning methods\cite{l3}, network embedding methods \cite{deepwalk,node2vec,struct2vec,line}, and graph convolution-based methods \cite{skipgnn,gcn,hogcn} for biomedical link prediction. We conduct our experiments on four publicly available datasets that represent various interaction in biomedical. ResMGCN outperforms the methods above both in accuracy and computation, demonstrating the superiority of ResNGCN mechanism.

We conduct case study on the PDI network to show that graph neural network using ResMGCN is able to learn meaningful informative entity representation of biomedical molecules.

In conclusion, our paper proposed residual graph convolution for predicting links between entities from the biomedical system, a novel graph convolution with leveraging low and high-order information in node representation learning, simple, fast, and effective. Our code and datasets are publicly available at https://github.com/Yonggie/ResMGCN. 

\section{Related work}
By utilizing molecular biomedical graphs, the goal of biomedical network prediction is to predict whether it has biomedical information correlation for two given entity nodes such as drug-drug, protein-protein, drug-gene, etc. There are three main flows of graph-based methods to achieve such goal: (1) \textbf{Heuristic learning} comes with the idea that it may exist a link if a pair of unconnected nodes are similar or close in the network. Methods such as \cite{pa,lp}, computes a similarity score by a certain expert-designed algorithm to represent how close the two nodes are. They are often limited to give strong constraints on the network and idealize the graph heavily to satisfy certain properties, which requires expert knowledge and departures from real-world scenarios. 
(2) \textbf{Direct graph embedding} such as \cite{line,node2vec,deepwalk} generates embeddings for nodes in the graph by absorbing the topology information. Even though these methods is intuitive and simple, they are limited to only model the topology information and lose the rich information of the nodes. (3) \textbf{Neural network based representation learning} utilizes graph convolution and learns graph entity informative representations and predicts node interactions by downstream predictor. In concrete, they learn node representation by aggregating the information from neighbour nodes (a.k.a. message passing\cite{hetero_survey}) and optimize the process by downstream task loss like link prediction in a convenient end2end manner. In GCN layer, spatially speaking, the node can only obtain its own first-order neighbor information in  message passing process and then update its own representation. It's the same with its neighbour nodes. In this case, the higher-order information from the far end can be indirectly obtained by stacking multiple graph convolutional layers, because its first-order neighbor also has obtained neighbor information of theirs. Details of this process are discussed in section \ref{gcn}.
\section{Preliminaries}
The graph in a biomedical system is defined as $\mathcal{G}=(\mathcal{V},\mathcal{E},\mathcal{X})$, where $\mathcal{V}$ is the set of nodes in biomedical entities such as gene, protein, drug, and $\mathcal{E}$ is the set of edges information denoting the interactions between the entities in biomedical system, and $\mathcal{X}\in R^{|\mathcal{V}|\times F}$ is the node feature representing  the information of each entity and can help graph convolution with additional information other than topology information, where $F$ is feature dimension.

Our goal in constructing biomedical information graph is to learn a predictor $P$ which is able to tell whether two arbitrary entities in a biomedical system interact with each other.
\subsection{Message Passing Framework}
\begin{figure}[t] 
       \centering      
       
         \begin{minipage}{0.25\textwidth}      
          \centering      
          \includegraphics[width = 1\textwidth]{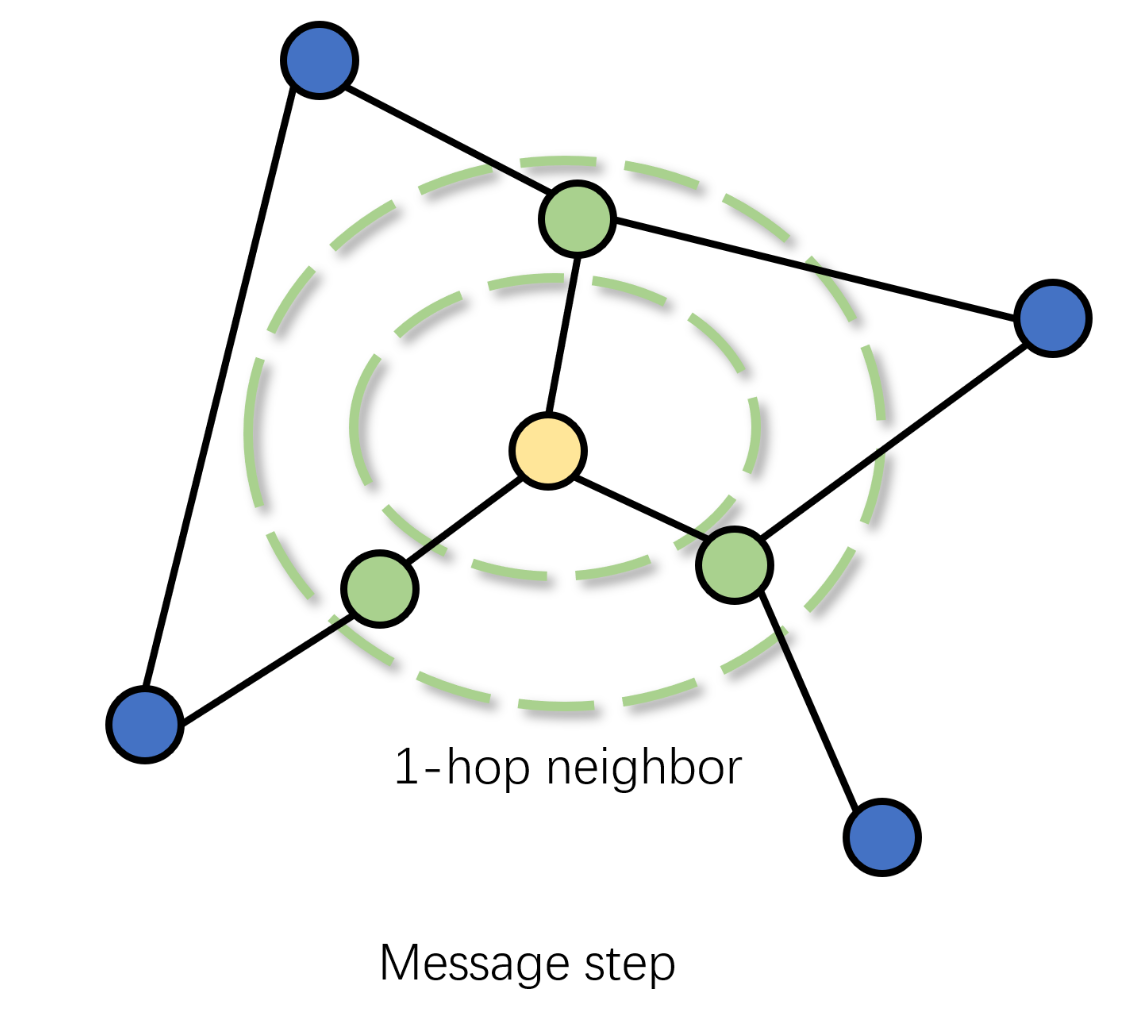}      
          \centerline{(a) Collect messages from the 1-hop neighbour.}
        \end{minipage}      
                   
          \begin{minipage}{0.25\textwidth}      
            \centering      
            \includegraphics[width = 1\textwidth]{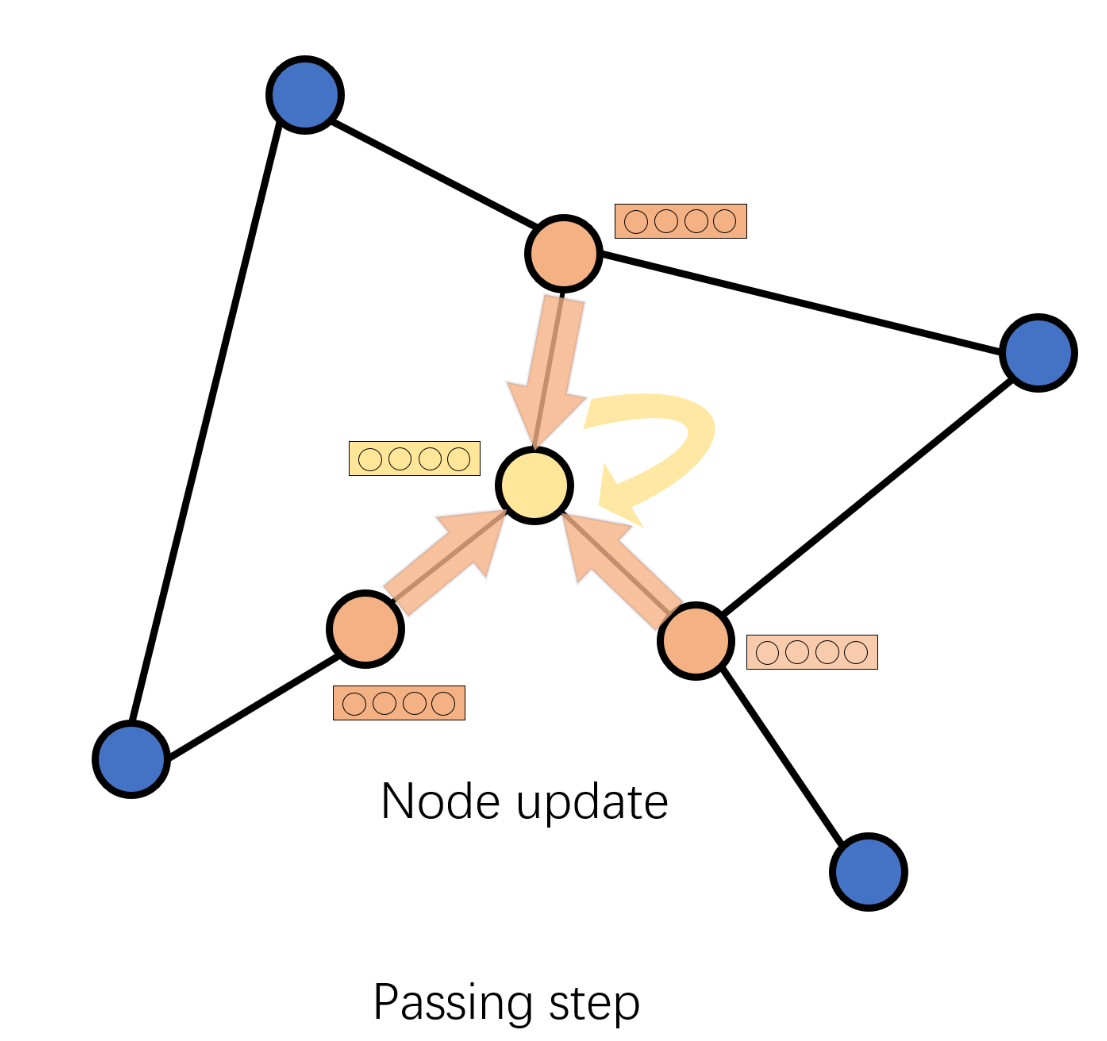}      
            \centerline{(b) Update the node representation accordingly.}
          \end{minipage}      
                
          \caption{Message passing framework illustration. Message step takes representations of multiple nearby neighbours to calculate the message the node needed for updating, formulated in Eq.\ref{eq:message}. Passing step updates the node representation according to the message provided by the first step and its own original representation, formulated in Eq.\ref{eq:passing}.}
          \label{fig:messagepassing}  
\end{figure}

In recent days, spatial graph convolution theory attracts more and more attentions after GCN is proposed. In message passing framework\cite{gmn}, as shown in Figure \ref{fig:messagepassing}, to learn an informative representation requires nodes interaction, and the interaction operation is divided into 2 steps: message and passing, where message is the various information to guide node representation learning and passing is an update process that utilizes the message into new representation. 

In most graph convolutions, the message $m_i$ in \textbf{message step} is calculated from nearby node representation, formulated as a function $f_m$ that takes representations of neighbouring nodes (1-hop neighbour) and calculates the message the central node $v_i$ needs: 
\begin{equation} 
\label{eq:message}
m_i=f_m(\{x_j\}), j \in \mathcal{N}_i
\end{equation}
where $\mathcal{N}_i$ is the index set of neighbourhood of node $v_i$.
In \textbf{passing step}, the node $v_i$ takes the message m$_i$ obtained in last step and its own representation $x_i$ to update the representation of node $v_i$, formulated as a update function $f_u$:\begin{equation} 
\label{eq:passing}
x_i'=f_u(m_i,x_i)
\end{equation}
Various of recently proposed graph convolutions apply different $f_m$ and $f_u$ as all kinds of variants of message passing framework.
\begin{figure}
	\centering
	\includegraphics[width=0.45\textwidth]{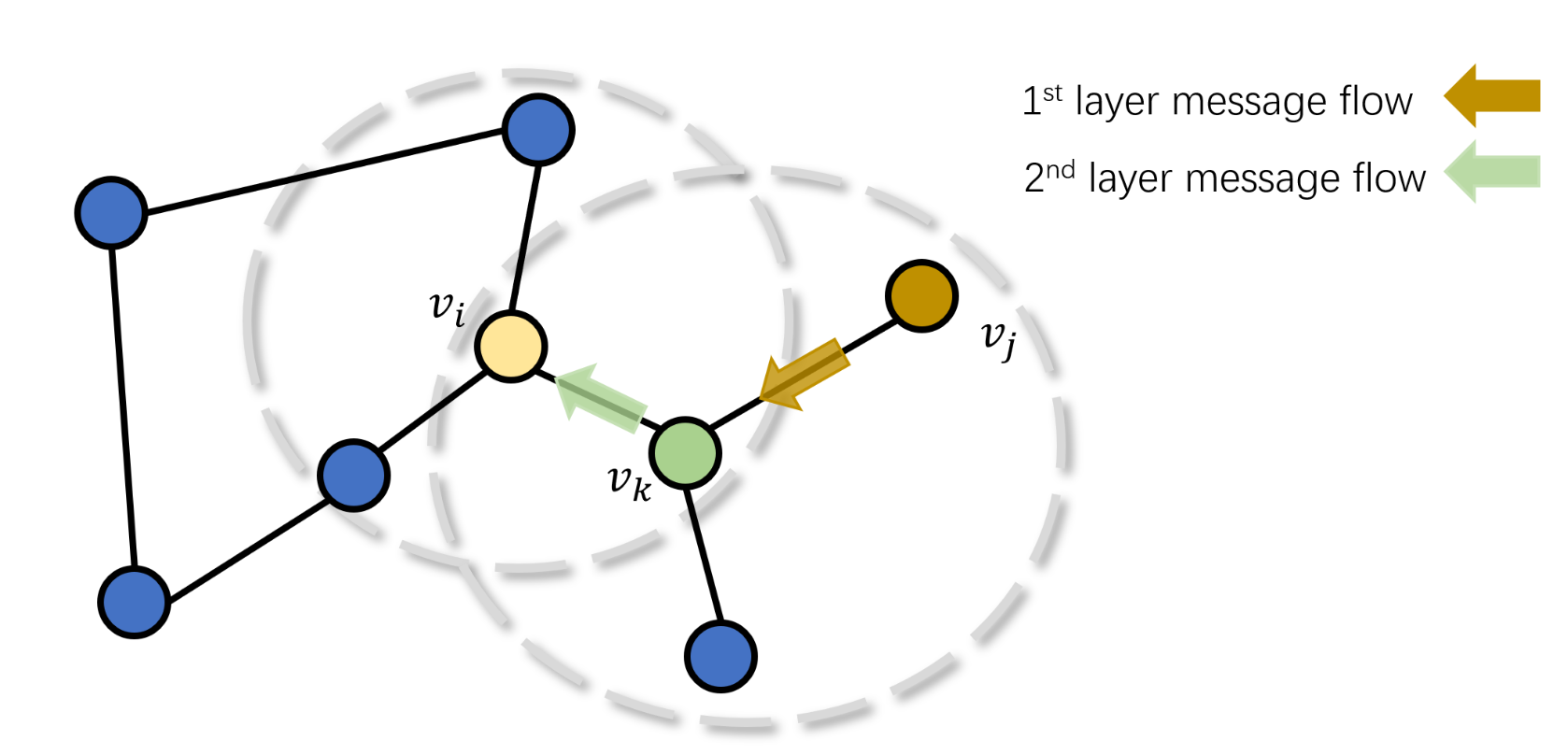}
	\caption{Illustration of multiplayer of message passing. Node $v_i$ is able to capture messages from node $v_j$ via node $v_k$ by stacking two layers of graph convolution.} 
  \label{fig:2hop}
\end{figure}

As an interesting case to note, nodes is able to capture k-hop (k $>$ 1) neighbours' message by stacking multiple graph convolution layers, in an indirect manner. As shown in Figure \ref{fig:2hop}, we take 2-layer message passing as an example, the first layer node only receives 1-hop messages of node $v_i$, while in the second layer node, however, node $v_i$ will be able to sense its 2-hop messages through the update of its 1-hop neighbour node $v_k$ indirectly, as the neighbour nodes $v_k$ is also updated with messages from its 1-hop neighbour such as node $v_j$, which is the 2-hop neighbour for node $v_i$.

\subsection{Graph Convolution(GCN)\label{gcn}}
Graph Convolution Network\cite{gcn} is a popular implement version of graph convolutions. Derived from spectral graph signal processing\cite{chebynet}\cite{neural_like}, it extends the convolution operation in computer vision or natural language processing to graph-structured data. In graph spectral space, the GCN is actually an ideal simplified version of ChebyNet\cite{chebynet} series, but in recent days people care more about what GCN processes the message in spatial space, as spatial space is much more intuitive, and thus developed a message passing system to describe different implementations of graph convolution. Concretely, GCN layer takes edge information as the adjacency matrix of graph $A$ and node feature matrix $X$ to update, where the element of adjacency  matrix $A$ is defined as:

\begin{equation}
A_{ij}=\begin{cases}
    1& if\ v_i\ connects\ v_j \\
    0& else
\end{cases}    
\end{equation}

In spectral space, GCN layer is denoted as:
\begin{equation} 
  X^{\prime}={f}(X, A)=\sigma\left(\hat{D}^{-\frac{1}{2}} \hat{A} \hat{D}^{-\frac{1}{2}} X W\right)
\end{equation}
where $D \in \mathbb{R}^{n \times n}$ is the degree matrix of the graph, a diagonal matrix with each degree of nodes on the diagonal, $n$ is the number of nodes, and $d$ is the feature dimension, and $\hat{A} \in \mathbb{R}^{n \times n} = A + I$ is the adjacency matrix with self-loops, and $X \in \mathbb{R}^{n\times d}$ is node feature matrix, $W \in \mathbb{R}^{d \times o}$ is a learnable linear transformation matrix of GCN to strengthen learning capability\cite{gcn}, $o$ is output dimension of this layer. 
\section{Proposed Residual GCN}
\subsection{Message Passing in GCN\label{mp_in_gcn}}
In the paper, GCN is proposed in a spectral form, on the other side, however, in spatial space, in an equivalent way\ref{proof},  we can also interpret GCN as follows: 
\begin{equation} 
x_i'=\sigma(\frac1{\sqrt{d_i}}(\frac1{\sqrt{d_i}}x_i+\sum_{j \in \mathcal{N}_i}\frac1{\sqrt{d_j}} x_j)W)
\end{equation}
where $x_i\in \mathbb{R}^{1 \times d}$ is the node representation of single node $v_i$.
To fit into the message passing framework, we separate the operation as follows. GCN sums up linear projected 1-hop  neighbour representation as message function, and updates the node by taking the mean value of the neighbouring and its own linear projected representation. GCN layer's operations are formulated as:
\begin{equation} 
m_i=f_m(\{x_j\})=\sum_{j \in \mathcal{N}_i} \frac1{\sqrt{d_j}}x_jW
\end{equation}
\begin{equation} 
x_i'=f_u(m_i,x_i)=\frac1{\sqrt{d_i}}({m_i+\frac1{\sqrt{d_i}}x_iW})
\end{equation}
The illustration of this process is illustrated in Figure \ref{fig:mp}.

\subsection{Residual Message GCN (ResMGCN)}
In biomedical data graphs, hierarchical connection correlations are crucial because new reactions or new drug molecular effects are often based on the progression of previous reaction information, which means information or message from recent lower order is extremely crucial, thus requires advanced information processing for messages from both last layer and current layer.

To address this problem, inspired by ResNet\cite{resnet}, we propose Residual Message GCN (ResMGCN), a simple but effective improvement for message fusion processing.

In the section \ref{mp_in_gcn}, the message $m_i$ denotes different information in the  different layer. For example, with $l$ denotes the layer index, for a 2-layer GCN, in the first layer $m^1$ is the message from 1-hop, while in the second layer, $m^2$ is the message from 2-hop neighbour via 1-hop neighbour, and so on. In order to jointly obtain hierarchical information and current information to guide node updates at the same time, we add a shortcut for message passing framework in different layers. Let $l$ the denote ResMGCN layer index, ResMGCN is formulated as:
\begin{equation} 
m_i^l=f_m(\{x_j\}), j \in \mathcal{N}_i
\end{equation}
\begin{equation} 
c_i^l=\Phi(m_i^l, m_i^{(l-1)})
\end{equation}
\begin{equation} 
x_i^l=f_u(c_i^l,x_i^{(l-1)})
\end{equation}
where $\Phi$ is a message fusion function to handle message from the current layer and last layer, to provide both lower and current information, and $l$ denotes layer index of ResMGCN, and $m_i^l$ denotes indirect distant message of $l$th layer, and $c_i^l$ denotes combined information from the different layer. The message from the previous layer flows into the current layer as message residue. The residual message and current message are fused by the fusion function $\Phi$ to guide node representation learning together.

\begin{figure}
	\centering
	\includegraphics[width=0.5\textwidth]{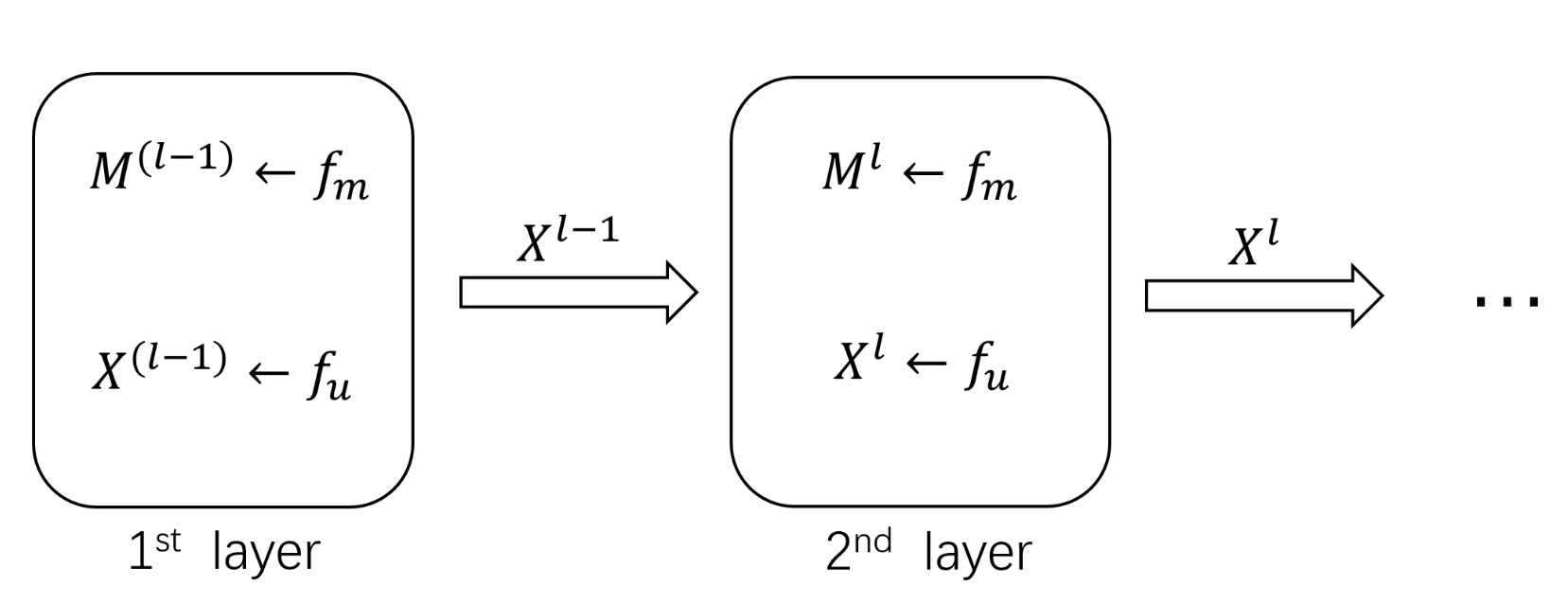}
	\caption{Illustration of message passing. The layer outputs the updated node representation $X$ for the next layer. The messages are left within the current layer.} 
  \label{fig:mp}
\end{figure}

\begin{figure}
	\centering
	\includegraphics[width=0.5\textwidth]{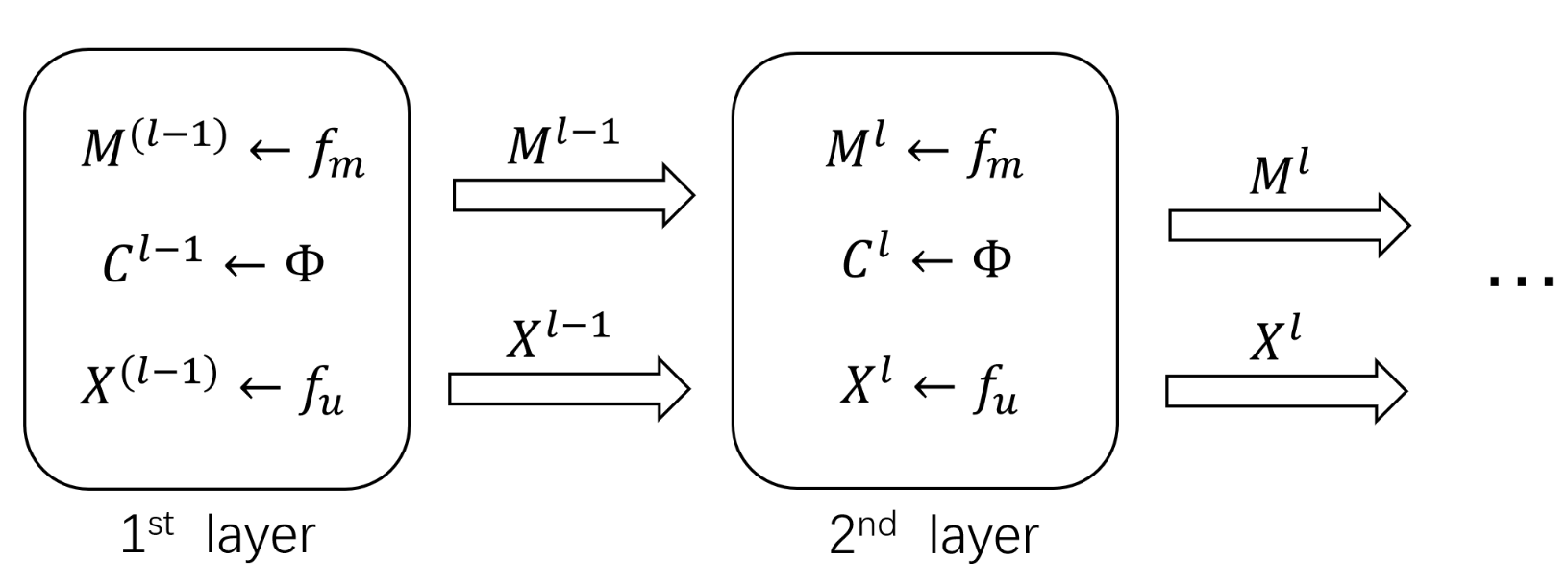}
	\caption{Illustration of ResMGCN. Besides of the updated representations, layer keeps the messages of current layer and pass the hierarchy information for next layer. Each layer has a information fusion function $\Phi$ to manage the information received for calculating the guided information $C$ for node updating.} 
  \label{fig:ResMGCN}
\end{figure}

In Figure \ref{fig:mp} and Figure \ref{fig:ResMGCN}, $M$ is the total messages of all nodes, and $C$ is the total fused messages of all nodes, and $l$ indicates the layer index, and $X$ denotes the node representation matrix. In the Figures, message or information from the lower hierarchy could flow better compared to the original, which provide fundamental interaction information for the current layer from the previous layer.
\section{ResMGCN for biomedical interaction prediction}
\subsection{Graph ResMGCN Feature Encoder}
We develop ResMGCN to fuse information from the different hierarchy of biomedical graphs. We conduct ResMGCN as graph feature encoder as a component in an end2end model and learn link prediction by linear predictor from biomedical graph representation. Specifically, the ResMGCN in end2end model is:
\begin{equation} 
m_i^l=f_m(\{x_j^{(l-1)}\})=\sum_{j \in \mathcal{N}_i} \frac1{\sqrt{d_j}}x_j^{(l-1)}W
\end{equation}
\begin{equation} 
c_i^l=\Phi(m_i^l, m_i^{(l-1)})=\frac12(m_i^l + m_i^{(l-1)})
\end{equation}
\begin{equation} 
x_i^l=f_u(c_i^l,x_i^l)=\sigma(\frac1{\sqrt{d_i}}({c_i^l+\frac1{\sqrt{d_i}}x_i^{(l-1)}}W))
\end{equation}

\subsection{Graph Predictor}
After feature encoding of the biomedical interaction graph, we introduce a simple linear predictor that takes the combination of feature from two entities and predicts the probability of interaction. For two embedded node representation $h_i$ and $h_j$, the predictor is formulated as:
$$p_{ij}=\sigma(\Psi(h_i,h_j)W_p+b)$$
where $p_{ij}$ is the probability of interaction between nodes $v_i$ and $v_j$, $W_p \in \mathbb{R}^{2d \times 1}$ is a learnable weight matrix of the predictor,$h_i \in \mathbb{R}^{1 \times d}$ and $h_j \in \mathbb{R}^{1 \times d}$ are node representation vectors of two arbitrary node $v_i$ and $v_j$, $d$ is feature dimension, $b$ is the bias of the predictor, $\sigma$ is a non-linear activation function, and $\Psi$ is a combination function that combines node feature $h_i$ and $h_j$ such as concatenation. We use vector concatenation as $\Psi$ function (same as SkipGNN) for a fair comparison with many baselines.

\subsection{Model Training and Optimization}
We apply binary cross entropy loss to optimize the model parameters:
\begin{equation} 
\mathcal{L}(v_i,v_j)=-((1-A_{ij})log(1-p_{ij})+A_{ij}log(p_{ij}))
\end{equation}
where $p_{ij}$ is the predicted interaction or reaction probability between node $v_i$ and $v_j$, and $A_{ij}$ denotes the label interaction (also the adjacency information) among nodes. Cross entropy loss could guide the model to predict a higher probability for ground truth interactions than that for non-interactions. For all nodes in training set, $\mathcal{L}$ is:
\begin{equation} 
\mathcal{L}=\sum_{(i,j)\in \mathcal{E}}\mathcal{L}(v_i,v_j)
\end{equation}
As an end2end model, the training information flows back from cross entropy loss to predictor and to feature encoder ResMGCN in the end during backpropagation.

\section{Experiment}
The biomedical interaction prediction problem is solved by a link prediction task on the interaction network. We conduct experiments on various interaction datasets and compare our ResMGCN with state-of-the-art models.

\subsubsection{Datasets}
We conduct link prediction experiments on four datasets in wide fields of bioinformatics, and all of them are publicly available. Details of the datasets are summarized in Table \ref{tb:dataset}.
\begin{itemize}
	\item BioSNAP-DTI\cite{dti}: DTI network contains 15,139 drug-target interactions between 5,018 drugs and 2,325 proteins.
    \item BioSNAP-DDI\cite{ddi}: DDI network contains 48,514 drug-drug interactions between 1,514 drugs extracted from drug labels and biomedical literature.
    \item HuRI-PPI\cite{ppi}: HI-III human PPI network contains 5,604 proteins and 23,322 interactions generated by multiple orthogonal high-throughput yeast two-hybrid screens.
    \item DisGeNET-GDI \cite{gdi}: GDI network consists of 81,746 interactions between 9,413 genes and 10,370 diseases curated from GWAS studies, animal models and scientific literature.
\end{itemize}

\begin{table*}[!htbp]
	\centering 
  \caption{Summary of the Datasets}
	\begin{tabular}
    {|ccccccc|} 
		\hline
		Dataset	   & \# Nodes   & \# Edges train  & \# Edges validate & \# Edges test & Edges Total & Avg. degree                     \\ 
		\hline
		DTI &5,018 drugs, 2,325 proteins & 10,597& 1,514 &3,028& 15,139 &4.12   \\ 
		
		DDI &1,514 drugs &33,960 &4,852& 9,702& 48,514 &64.09\\
PPI &5,604 proteins &16,326 &2,332& 4,664& 23,322 &8.32\\
GDI &9,413 genes, 10,370 diseases& 57,222 &8,175& 16,349& 81,746& 8.26\\
\hline
	\end{tabular}
	
	\label{tb:dataset}
\end{table*}

\subsection{Baselines}
We compare ResMGCN with seven powerful predictors of molecular interactions from network science and graph machine-learning fields. For direct embedding models, we use three direct network embedding methods: DeepWalk\cite{deepwalk}, node2vec\cite{node2vec}, and we also include struct2vec\cite{struct2vec}. struct2vec and node2vec are conceptually distinct by leveraging local network structural information, while Deepwalk uses random walks to learn embeddings for nodes in the network. Moreover, we compare with five graph neural networks: VGAE\cite{vgae},GCN\cite{gcn},GAT\cite{gat}, GIN\cite{gin},JK-Net\cite{jknet}, MixHop\cite{mixhop} and HOGCN\cite{hogcn}. These methods utilize graph convolution to extract the informative representations of biomedical entities, with the same input as ResMGCN. For heuristic learning, we consider
Spectral Clustering\cite{sc} and L3\cite{l3} heuristic. 
\subsection{Metrics}
We use the same evaluations for all baselines. Datasets are divided into training, validation, and test sets in the ratio of 7:1:2. We use AUROC, which is the area under the receiver operating characteristic curve, and AUPRC, which is the area under the Precision-Recall Curve as two evaluation indicators.
\subsection{Experimental Settings}
For a fair comparison, we provide node initial representation as one hot initialization and linear link predictor for SkipGNN, HOGCN, and ResMGCN. SkipGNN implemented by the paper author used one hot initialization, and HOGCN implemented by the paper author used node2vec representation initialization to offer informative embedding in the beginning and did not provide the initial embedding in the Github project. According to the variable-controlling method, we initialize the node representation with one hot and only change the graph feature encoder in these three methods to reveal the true comparison of these methods.
\subsection{Results}
In this section, we demonstrate the performance of baselines and ResMGCN, and analysis details about the results. The summary of performance is in Table \ref{tb:performace}. We referred to the statistics from the previous paper\cite{hogcn,skipgnn} and conduct a fair implementation of HOGCN experiment.

ResMGCN shows very competitive performance compared with all baselines including state-of-the-art methods both on AUPRC and AUROC on all four biomedical datasets. ResMGCN outperformed all baselines on DDI and GDI datasets. In DDI dataset, ResMGCN achieved around 7\% higher and 5\% higher than previous sota SkipGNN in  AUPRC and AUROC respectively. In GDI dataset, ResMGCN achieved around 1\% higher and 0.4\% higher than previous sota HOGCN in AUPRC and AUROC respectively. In these four datasets, DDI has the highest node average degree 64.09, whereas other datasets have only less than 10 average degree. ResMGCN has the best performance in DDI dataset, proving that ResMGCN is good at handling different hierarchies and dense topology information. ResMGCN has huge advantages on both data scale and training cost, even though this is not so revealed in rather in datasets on smaller scales. Dataset GDI has the greatest amount of nodes and edges (around 20,000), and ResMGCN outperformed the baselines, together with later discussion\ref{trainingcost} on training cost analysis, demonstrating the ResMGCN has good capability of dealing with large dataset. 


\begin{table}[!htbp]
    \renewcommand\arraystretch{1.2}
	\centering 
  \caption{Average AUPRC and AUROC With One Standard Deviation
on Biomedical Interaction Prediction}
	\begin{tabular}
    {|ccccc|} 
		\hline
		Dataset	   & Method  & AUPRC  & AUROC & Rank\\ 
		\hline
		DTI &DeepWalk &0.753 ± 0.008& 0.735 ±  0.009&\\
            &node2vec &0.771 ±  0.005 &0.720 ±  0.010&\\
            &struct2vec &0.677  ±  0.007& 0.656  ±  0.010&\\
            &SC &0.818  ±  0.007 &0.743  ±  0.008& \\
            &L3 &0.891 ±  0.004 &0.793  ± 0.006&\\
            &VGAE &0.853 ±  0.010& 0.800 ±  0.010&\\
            &GCN &0.904  ± 0.011 &0.899  ± 0.010&\\
            &GIN&0.922  ±  0.004 &0.907  ±  0.006&4\\
            &JK-Net& 0.921  ±  0.006& 0.907  ±  0.008& \\
            &MixHop&0.921  ±  0.006 &0.920  ±  0.004&2 \\
            &SkipGNN &0.928 ±  0.006& 0.922 ±  0.004&1\\
            &HOGCN &0.929 ±  0.001& 0.919  ± 0.001& 3\\ 
            &ResMGCN &0.918 ± 0.001 &0.901 ± 0.010&5 \\
		\hline
		
		DDI &DeepWalk &0.698 ±  0.012 &0.712  ± 0.009&\\
            &node2vec &0.801  ± 0.004 &0.809 ±  0.002&\\
            &struct2vec&0.643  ±  0.012 &0.654  ±  0.007& \\
            &SC &0.749  ±  0.009 &0.816  ±  0.006& \\
            &L3 &0.860  ± 0.004 &0.869  ± 0.003&\\
            &VGAE &0.844  ± 0.076 &0.878  ± 0.008&\\
            &GCN &0.856  ± 0.005 &0.875  ± 0.004&\\
            &GIN &0.856  ±  0.005 &0.876  ±  0.003& \\
            &JK-Net &0.870  ±  0.009 &0.885  ±  0.005&3 \\
            &MixHop &0.861  ±  0.006 &0.879  ±  0.004&4\\
            &SkipGNN &0.866  ± 0.006 &0.886  ± 0.003&2\\
            &HOGCN &0.8501 ±  0.003 &0.8620 ±  0.002&\\
            &ResMGCN &0.9317 ± 0.001 &0.9360 ± 0.001&1\\
		\hline
        PPI &DeepWalk &0.715  ± 0.008 &0.706  ± 0.005&\\
            &node2vec &0.773 ±  0.010 &0.766 ±  0.005&\\
            &struct2vec &0.875  ±  0.004& 0.868  ±  0.006&\\
            &SC &0.897  ±  0.003 &0.859  ±  0.003 &\\
            &L3 &0.899  ± 0.003 &0.861  ± 0.003&\\
            &VGAE &0.875  ± 0.004 &0.844  ± 0.006&\\
            &GCN &0.909  ± 0.002 &0.907  ± 0.006&\\
            &GIN &0.907  ±  0.004 &0.897  ±  0.006& \\
            &JK-Net &0.912  ±  0.003 &0.901  ±  0.005&3 \\
            &MixHop &0.909  ±  0.004 &0.913  ±  0.003&\\
            &SkipGNN &0.921  ± 0.003 &0.917  ± 0.004&1\\
            &HOGCN & 0.9155 ± 0.002 &0.9104  ± 0.001&2\\
            &ResMGCN &0.9104 ± 0.001 &0.8940 ± 0.001&4\\
		\hline
        GDI &DeepWalk &0.827  ± 0.007 &0.832 ±  0.003&\\
            &node2vec &0.828  ± 0.006 &0.834 ±  0.003&\\
            &struct2vec &0.910  ±  0.006& 0.909  ±  0.005&\\
            &SC &0.905  ±  0.002 &0.863  ±  0.003& \\
            &L3 &0.899  ± 0.001 &0.832  ± 0.001&\\
            &VGAE &0.902  ± 0.006 &0.873  ± 0.009&\\
            &GCN &0.909  ± 0.002 &0.906  ± 0.006&\\
            &GIN &0.916  ±  0.004& 0.900  ±  0.005& \\
            &JK-Net& 0.891  ±  0.049 &0.898  ±  0.002& \\
            &MixHop& 0.912  ±  0.005 &0.916  ±  0.004&3 \\
            &SkipGNN& 0.915  ± 0.003& 0.912  ± 0.004&4\\
            &HOGCN &0.924  ± 0.001 &0.921  ± 0.001&2\\
            &ResMGCN &0.935 ± 0.001 &0.925 ± 0.001&1\\
        \hline
	\end{tabular}
	
	\label{tb:performace}
\end{table}

\subsection{ResMGCN is \textbf{simple and much more effective}\label{trainingcost}}
In this section, we compare ResMGCN effectiveness with state-of-the-art methods in recent years. We conduct model training on the same server executing one single task at a time with RTX 3090s. The time cost of hhe average epoch and training process with their best epoch number, which is the epoch round that would give the model the best performance on biomedical indicators, are listed below in Table \ref{tb:time}. As time cost is also related to implementation acceleration, we clarify that these methods are all based on basic matrix multiplication provided by Pytorch.
\begin{table}[!htbp]
    \renewcommand\arraystretch{1.2}
	\centering 
  \caption{Summary of the effectiveness of related methods}
	\begin{tabular}
    {|cccc|} 
		\hline
		Dataset	   & Method  & Epoch  & Training \\ 
		\hline
		DTI 
            &SkipGNN& 14.8 & 241.8\\
            &HOGCN &4.70 &119.6\\
            &ResMGCN &2.6 & 105.7\\
        \hline
		
		DDI 
            &SkipGNN& 55.04& 749.4\\
            &HOGCN & 12.5 &503.1\\
            &ResMGCN & 6.3& 81.7\\
        \hline
        PPI 
            &SkipGNN& 20.6& 310.2\\
            &HOGCN &6.2 &161.1\\
            &ResMGCN & 3.5 & 55.5\\
        \hline
        GDI 
            &SkipGNN& 180.9 & 903.6\\
            &HOGCN &20.3 & 320.0\\
            &ResMGCN & 9.8 & 50.2\\
        \hline
	\end{tabular}
	
	\label{tb:time}
\end{table}

We obviously see our proposed ResMGCN reached state-of-the-art effectiveness compared with all related methods. (1) SkipGNN reconstructed a copy of the original graph with 1-hop edges removed and 2-hop edges connected to obtain higher-order information and then aggregates features from the two graphs, which requires large computation for graph preprocessing and doubled the memory and time cost. It is too expensive to conduct such operation in a real large scale scenario. Despite of the performance superiority of ResMGCN, the time cost of ResMGCN is \textbf{significantly} lower than SkipGNN, nearly \textbf{one-tenth} of SkipGNN in terms of single epoch cost and up to \textbf{one eighteenth} ($\frac1{18}$) of that in terms of single epoch cost. ResMGCN converges in a significantly fast and effective manner with a superb performance in many of the datasets. (2) HOGCN is a method that stacks neighbour messages from 0-hop to up to k-hop in one layer as information fusion, where it requires multiple matrix multiplications in one single layer, which could cost a huge amount of time. To gain a fair comparison, we implemented experiments on DDI dataset in the same settings with HOGCN\cite{hogcn} in sparse adjacency matrix and achieved sota effectiveness, but our model could \textbf{reach 5.2s per epoch} on this dataset if we use dense adjacency matrix. Despite of the performance superiority of ResMGCN, ResMGCN could reach nearly \textbf{half} the time of HOGCN in terms of a single epoch and up to \textbf{one sixth} ($\frac16$) the time of HOGCN in terms of whole training cost. Other than that, HOGCN\cite{hogcn} needs human hyperparameter k to preprocess information in the beginning, while ResMGCN is a simple and fully end2end convolution that doesn't require any intervention. 

The larger the dataset scale, the greater advantages of ResMGCN. The advantage of ResMGCN is more obvious when it is applied in large datasets such as GDI dataset. We see greater epoch cost margin compared with PPI whose node number is 1,514 and GDI whose node number is around 20,000.

As mentioned before, we conduct a fair comparison, we did not apply bilinear edge reconstruction as HOGCN did, and did not use any trick for higher performance, which means this is not the end of ResMGCN's potential, and further exploration works are expected.


\section{Conclusion}
In this paper, we proposed a novel graph convolution ResMGCN, a new message passing scheme to fuse information from different hierarchies to guide more meaningful graph node representations. We conduct solid experiments on four publicly available datasets on interaction prediction and compare ResMGCN with state-of-the-art baselines. ResMGCN outperforms all baselines with clear improvement on interaction prediction task on the biomedical information graph, as ResMGCN learns informative representation through interacted messages from various orders of neighbour. Moreover, the proposed shortcut scheme enables ResMGCN to process significantly fast and effective feature encoding both on computation and storage compared with the methods which try to alleviate the similar issue. The experiments of biomedical interaction prediction and effectiveness comparison demonstrates ResMGCN's superiority and great potential in other related areas.


%

\appendices
\section{Proof of the Equivalence of spatial and spectral convolution of GCN\label{proof}}
\begin{equation} 
  X^{\prime}={f}(X, A)=\sigma\left(\hat{D}^{-\frac{1}{2}} \hat{A} \hat{D}^{-\frac{1}{2}} X W\right)
\end{equation}
where $D \in \mathbb{R}^{n \times n}$ is the degree matrix of the graph, $A \in \mathbb{R}^{n \times n}$ is the adjacency matrix of the graph, $X \in \mathbb{R}^{n \times d}$ is the representation matrix of all nodes in the graph. 
Taking a graph composed of three nodes as an example, we explicitly demonstrate the every element of $D$,$A$ and $X$ matrices. These matrices are denoted as:

\begin{small}
$$
\setlength{\arraycolsep}{0.7pt}
 \left[\!
 \begin{array}{c}
     x'_{1}   \\
     x'_{2}   \\
     x'_{3}   
 \end{array}\!
 \right]\!=\!\sigma(
\left[
 \begin{array}{ccc}
     d_{1}^{-\frac12} & 0 & 0 \\
     0 & d_{2}^{-\frac12} & 0 \\
     0 & 0 & d_{3}^{-\frac12} 
 \end{array}\!
 \right] 
\left[
 \begin{array}{ccc}
     a_{11} & a_{12} & a_{13} \\
     a_{21} & a_{22} & a_{23} \\
     a_{31} & a_{32} & a_{33} 
 \end{array}\!
 \right]\!\left[
 \begin{array}{ccc}
     d_{1}^{-\frac12} & 0 & 0 \\
     0 & d_{2}^{-\frac12} & 0 \\
     0 & 0 & d_{3}^{-\frac12} 
 \end{array}
 \right] 
 \left[
 \begin{array}{c}
     x_{1}   \\
     x_{2}   \\
     x_{3}   
 \end{array}\!
 \right] 
W)
$$
\end{small}


where $d_i$ is a scalar, the degree of node $v_i$, $a_{ij}$ is a scalar, the element of adjacency matrix $A$, $x_i \in \mathbb{R}^{1 \times d}$ is the representation of single node $v_i$, and $x_i' \in \mathbb{R}^{1 \times d}$ is the updated representation of node $v_i$, the output of this layer.

\begin{small}
$$
\setlength{\arraycolsep}{0.5pt}
 \left[
 \begin{array}{c}
     x'_{1}   \\
     x'_{2}   \\
     x'_{3}   
 \end{array}
 \right] 
=\sigma(
\left[
 \begin{array}{ccc}
     d_{1}^{-\frac12}\! & 0 & 0 \\
     0 & d_{2}^{-\frac12} & 0 \\
     0 & 0 & d_{3}^{-\frac12} 
 \end{array}
 \right] 
\left[
 \begin{array}{ccc}
     a_{11} & a_{12} & a_{13} \\
     a_{21} & a_{22} & a_{23} \\
     a_{31} & a_{32} & a_{33} 
 \end{array}
 \right]
 \left[
 \begin{array}{ccc}
     d_{1}^{-\frac12} & 0 & 0 \\
     0 & d_{2}^{-\frac12} & 0 \\
     0 & 0 & d_{3}^{-\frac12} 
 \end{array}
 \right] 
 XW)
$$
\end{small}

To reveal the updating of representation $x_i'$ of node $v_i$, we let $Z=D^{-{\frac12}}AD^{-{\frac12}}$. $Z$ is:

$$
Z=\left[
    \begin{array}{ccc}
    d_1^{-\frac12}d_1^{-\frac12}a_{11}&d_1^{-\frac12}d_2^{-\frac12}a_{12}&d_1^{-\frac12}d_3^{-\frac12}a_{13}\\
    d_2^{-\frac12}d_1^{-\frac12}a_{21}&d_2^{-\frac12}d_2^{-\frac12}a_{22}&d_2^{-\frac12}d_3^{-\frac12}a_{23}\\
    d_1^{-\frac12}d_3^{-\frac12}a_{31}&d_3^{-\frac12}d_2^{-\frac12}a_{32}&d_3^{-\frac12}d_3^{-\frac12}a_{33}
    \end{array}
    \right]   
$$

We multiply Z with the representation matrix in vector format $X=(x_1^T,x_2^T,x_3^T)$:

$$
ZX=
\left[
    \begin{array}{ccc}
    d_1^{-\frac12}d_1^{-\frac12}a_{11}&d_1^{-\frac12}d_2^{-\frac12}a_{12}&d_1^{-\frac12}d_3^{-\frac12}a_{13}\\
    d_2^{-\frac12}d_1^{-\frac12}a_{21}&d_2^{-\frac12}d_2^{-\frac12}a_{22}&d_2^{-\frac12}d_3^{-\frac12}a_{23}\\
    d_1^{-\frac12}d_3^{-\frac12}a_{31}&d_3^{-\frac12}d_2^{-\frac12}a_{32}&d_3^{-\frac12}d_3^{-\frac12}a_{33}
    \end{array}
\right]\!
\left[\!
 \begin{array}{c}
     x_{1}   \\
     x_{2}   \\
     x_{3}   
 \end{array}\!
 \right] 
$$

$$
\setlength{\arraycolsep}{0.7pt}
ZX=\left[
    \begin{array}{ccc}
    d_1^{-\frac12}(d_1^{-\frac12}a_{11}x_1+d_2^{-\frac12}a_{12}x_2+d_3^{-\frac12}a_{13}x_3)\\
    d_2^{-\frac12}(d_1^{-\frac12}a_{21}x_1+d_2^{-\frac12}a_{22}x_2+d_3^{-\frac12}a_{23}x_3)\\
    d_3^{-\frac12}(d_1^{-\frac12}a_{31}x_1+d_2^{-\frac12}a_{32}x_2+d_3^{-\frac12}a_{33}x_3
    )\end{array}
    \right]
$$

As $X'=\sigma(ZXW)$, finally, the node updating of nodes $v_i$ is:
\begin{small}
$$
\left[\!
 \begin{array}{c}
     x'_{1}   \\
     x'_{2}   \\
     x'_{3}   
 \end{array}\!
 \right] 
= \sigma(\left[
    \begin{array}{ccc}
    d_1^{-\frac12}(d_1^{-\frac12}a_{11}x_1+d_2^{-\frac12}a_{12}x_2+d_3^{-\frac12}a_{13}x_3)\\
    d_2^{-\frac12}(d_1^{-\frac12}a_{21}x_1+d_2^{-\frac12}a_{22}x_2+d_3^{-\frac12}a_{23}x_3)\\
    d_3^{-\frac12}(d_1^{-\frac12}a_{31}x_1+d_2^{-\frac12}a_{32}x_2+d_3^{-\frac12}a_{33}x_3
    \end{array}
    \right]W)
$$
\end{small}

Taking $x_1'$ as an example, the updating process is:

$$
x_1'=\sigma(d_1^{-\frac12}(d_1^{-\frac12}a_{11}x_1+d_2^{-\frac12}a_{12}x_2+d_3^{-\frac12}a_{13}x_3)W)
$$

This equation demonstrates an equivalent spatial manner that every node takes its 1-hop neighbour in spatial graph and aggregates the messages to updates its own representation. Thus, in the aspect of spectral domain, we can formulate the GCN  process on one node as:
\begin{equation}
    x_i'=\sigma(\frac1{\sqrt{d_i}}(\frac1{\sqrt{d_i}}x_i+\sum_{j \in \mathcal{N}_i}\frac1{\sqrt{d_j}} x_j)W)
\end{equation}
where $\mathcal{N}_i$ is the index set of neighbourhood of node $v_i$.

\ifCLASSOPTIONcaptionsoff
  \newpage
\fi

\bibliographystyle{IEEEtran}
\bibliography{my_ref}

\end{document}